\documentclass{article} % For LaTeX2e
\usepackage{iclr2023,times}
\usepackage{amsmath,amsfonts,bm}

\def\eqref#1{equation~\ref{#1}}
\def\1{\bm{1}}

\DeclareMathAlphabet{\mathsfit}{\encodingdefault}{\sfdefault}{m}{sl}
\SetMathAlphabet{\mathsfit}{bold}{\encodingdefault}{\sfdefault}{bx}{n}

\usepackage{mathtools}
\usepackage[mathscr]{eucal}
\usepackage{hyperref}
\usepackage{url}
\usepackage{graphicx}
\usepackage{glossaries}
\robustify{\gls}
\usepackage{booktabs}
\usepackage{multicol}
\usepackage{multirow}
\usepackage{longtable}
\usepackage{tabularx}
\usepackage{caption}
\usepackage{subcaption}
\usepackage{paralist}
\usepackage{colortbl}
\usepackage{algorithm}
\usepackage{algorithmicx}
\usepackage[noend]{algpseudocode}
\usepackage[capitalise]{cleveref}%it must be the last one to load!

\graphicspath{%
  {figures/}%
  {figures/density/}%
  {figures/erroneous_graph/}%
  {figures/mechanisms/}%
  {figures/nb_data/}%
  {figures/nb_variables/}%
  {figures/outliers/}%
  {figures/scm_noise/}%
  {figures/weight_threshold/}%  
}

\hypersetup
{
  colorlinks=true,
  linktocpage=false,
  linkcolor=black,
  urlcolor=blue,
  citecolor=black,%red, %teal
  pdfstartview=FitH,
  filecolor=black,%DarkScarletRed,
  pdftitle={},
  pdfauthor={},
  pdfcreator={},
  pdfsubject={},
  pdfkeywords={},%
}

\newcommand{\ie}{i.e.,}
\newcommand{\eg}{e.g.,}

\title{A Guide for Practical Use of ADMG Causal Data Augmentation}

\author{Audrey Poinsot\(^{1,2}\), Alessandro Leite\(^{1}\) \\
TAU, INRIA, LISN, Paris-Saclay University\(^{1}\)\\ 
Ekimetrics\(^{2}\)\\
\texttt{audrey.poinsot@inria.fr}, \; \texttt{alessandro.leite@inria.fr} \\
}

\newacronym{ml}{ML}{machine learning}
\newacronym{ai}{AI}{artificial intelligence}
\newacronym{cda}{CDA}{causal data augmentation}
\newacronym{cm}{CM}{causal model}
\newacronym{cg}{CG}{causal graph}
\newacronym{da}{DA}{data augmentation}
\newacronym{cv}{CV}{computer vision}
\newacronym{nlp}{NLP}{natural language processing}
\newacronym{admg}{ADMG}{acyclic-directed mixed graph}
\newacronym{dag}{DAG}{directed acyclic graph}
\newacronym{scm}{SCM}{structural causal model}
\newacronym{gmm}{GMM}{gaussian mixture model}
\newacronym{mape}{MAPE}{mean absolute percentage error}
\newacronym{xgb}{XGBoost}{eXtreme Gradient Boosting}
\newacronym{mcts}{MCTS}{Monte Carlo Tree Search}
\newacronym{mdl}{MDF}{minimum description length}
\newacronym{ca}{CD}{CausalDA}

\iclrfinalcopy % Uncomment for camera-ready version, but NOT for submission.
\begin{document}

\maketitle

\begin{abstract}\label{sec:abstract}%
Data augmentation is essential when applying~\gls{ml} in small-data regimes. It generates new samples following the observed data distribution while increasing their diversity and variability to help researchers and practitioners improve their models' robustness and, thus, deploy them in the real world. Nevertheless, its usage in tabular data still needs to be improved, as prior knowledge about the underlying data mechanism is seldom considered, limiting the fidelity and diversity of the generated data. Causal data augmentation strategies have been pointed out as a solution to handle these challenges by relying on conditional independence encoded in a causal graph. In this context, this paper experimentally analyzed the~\gls{admg} causal augmentation method considering different settings to support researchers and practitioners in understanding under which conditions prior knowledge helps generate new data points and, consequently, enhances the robustness of their models. The results highlighted that the studied method \begin{inparaenum}[(a)] \item is independent of the underlying model mechanism, \item requires a minimal number of observations that may be challenging in a small-data regime to improve an~\gls{ml} model's accuracy, \item propagates outliers to the augmented set degrading the performance of the model, and \item is sensitive to its hyperparameter's value\end{inparaenum}\@.%
\end{abstract}%

\section{Introduction}\label{sec:introduction}
\Acrfull{ml} models require quality data to be able{}
% \Gls{ml} models depend on the characteristics of the training data 
to discover helpful information, perform well on unseen data, and be robust to environmental changes. Although some models can handle noisy and high-dimensional datasets, their usage in high-stake small-data regimes is usually challenging. In this case, one can use~\acrlong{da} techniques to deal with the lack of training data to improve models' performance and limit overfitting by artificially increasing the number of samples and the diversity of the training set~\citep{van:01}. 
% 
% \Acrlong{da} methods aim to generate samples following the observed data distribution while increasing the diversity and variability inside the dataset~\citep{van:01}.
They have been successfully used in~\gls{cv}~\citep{zhong:20,hendrycks:21} and~\gls{nlp}~\citep{xie:20, hao:23} tasks, by providing model regularization during training and consequently, helping reducing overfitting. Nonetheless, these techniques cannot be easily extended to tabular or time series data~\citep{talavera:22}. Likewise, they usually focus on increasing samples' diversity or variability~\citep{qingsong:21}\@ and rarely both.

Knowing the underlying causal mechanism may help data augmentation techniques handle these issues by taking advantage of partial knowledge encoded in a~\gls{cg}. %A~\acrlong{cg}, in this case, represents prior knowledge. 
Thus, once one has been built, we can use it to infer the conditional independence relations that a data distribution should satisfy. %Moreover, \acrlongpl{cm} allow one to model distribution shifts, handle spurious correlations through the notion of intervention (\ie{} \emph{what if}), and address the question of what interventions can result in a given effect~\citep{pearl:09,pearl:19}. In other words, they encode causal information describing how data distributions change under an intervention\@.
As a result, one can combine data from an interventional distribution with augmented and observed ones~\citep{ilse:21} to improve both the diversity and variability of a dataset, hoping to improve the robustness of an~\gls{ml} model. Such a strategy can be implemented by following a causal boosting procedure~\citep{little:19} or exploring prior knowledge of conditional independence encoded in a causal graph~\citep{teshima:21}. The former generates new samples by weighting the data coming from intervention distributions. %with a focus on performance degradation
In contrast, the latter generates new data samples by simultaneously considering all possible resampling paths from the conditional empiric distribution of each variable assuming the existence of an~\acrfull{admg}~\citep{richardson:03}. %with a focus on performance improvement due to the knowledge given by the~\acrlong{cg}.

In this context, this paper experimentally\footnote{The code is available at~\url{github.com/audreypoinsot/admg_data_augmentation}} assesses the characteristics of the~\gls{admg} data augmentation method~\citep{teshima:21}~(\Cref{sec:causalda}) under the fidelity, diversity, and generalization~\citep{alaa:22} perspective in a small-data regime configuration, considering different problem's properties, and with the presence of noisy data~(\Cref{sec:setup}). The goal is to understand under which conditions this method can help practitioners increase the robustness and deal with overfitting of their~\gls{ml} models by augmenting their datasets using prior knowledge encoded in a~\acrlong{cg}. Another objective comprises understanding under which conditions an inadequate parametrization setting can lead to unexpected results; \ie{} performance degradation\@.

\section{ADMG Data Augmentation}\label{sec:causalda}

In this section, the~\gls{admg} causal data augmentation method~\citep{teshima:21} is presented. From now on, we refer to this method as~\acrlong{ca}.

Let us assume we want to train a model $f$ using the loss $L$ on the dataset $D=(D_{train}, D_{test})$ composed of a training  and a testing set with $d$ dimensions. Let us assume a known~\gls{admg} causal graph \(\mathcal{G}\) linking the $d$ variables ordered according to the topological order induced by the graph $\mathcal{G}$.

Let us use the following notations:

\begin{itemize}
    \item $n=|D_{train}|$ the number of training data
    \item $X_k$ the $k^{th}$ data point of the training set $D_{train}=\{X_i\}_{i\in [1,n]}$
    \item $X_k^j$ the value taken by the $j^{th}$ variable of the $k^{th}$ training point, $X_k = \{X_k^1, ..., X_k^d\}$
    \item $X_k^J$ with $J$ a set of variables, the value taken by the $J$ variables of the $k^{th}$ training point
    \item $D_{aug}$ the augmented dataset using $D_{train}$
    \item $Z_i$ the $i^{th}$ augmented data point from $D_{aug}$
    \item $a(j)$ the ancestors of the variable $j$ in the causal graph $\mathcal{G}$
\end{itemize}

$D_{aug}$ is built as the cartesian product of all the observed variables in the training set:
\begin{center}
    $D_{aug}=\{Z_i\}_{i\in [1,n^d]}$, \;\; $Z_i=\{X_{i_1}^1, ..., X_{i_j}^j, ..., X_{i_d}^d\}$
\end{center}
with $X_{i_j}$ the data point used to copy its value of the $j^{th}$ variable to use for the augmented point $Z_i$.

Each $Z_i$ is associated with a weight $w_i$ which could be interpreted as a probability of existence for the augmented point $Z_i$. Indeed, $w_i$ measures the probability of the variables values of the augmented point $Z_i$ given variables ancestor values. Probabilities are estimated with Kernels, $K^j$ denoting the kernel used to estimate the probability of the $j^{th}$ variable given its ancestors.

\begin{equation}\label{eq:weight}
    w_i = \prod_{j=1}^d w_i^j = \prod_{j=1}^d \frac{K^j(Z_i^{a(j)}-X_{i_j}^{a(j)})}{\sum_{k=1}^n K^j(Z_i^{a(j)}-X_k^{a(j)})}
\end{equation}

Finally, a model $f$ is trained on the augmented set using a weighted loss: 

\begin{equation}\label{eq:loss}
    L_{aug}(f) = \sum_{i\in [1,n^d]}w_iL(f,Z_i)
\end{equation}

In practice, the weights are computed recursively through~\cref{alg:causalDA}. In order to reduce memory and computational cost, the method enables us to choose a probability threshold $\theta \in [0,1[$ to early stop the computation of a weight (and the associated augmented point) as soon as its current value is lower than $\theta$.

\begin{algorithm}
\caption{\acrlong{ca} algorithm}\label{alg:causalDA}
\textbf{Input:} $D_{train}=\{X_k\}_{k\in [1,n]}$, $\;\mathcal{G}$, $\;\theta$, $\;L$, $\;\{K^j\}_{j\in [1,d]}$ \Comment{assuming that the variables in the training set and kernel functions are ordered according to the topological order of the graph \(\mathcal{G}\)}
\begin{algorithmic}
% \State \textbf{Order} the $d$ variables following the topological order given by $\mathcal{G}$
\vspace{2pt}
\State $W_{aug} \gets \{\frac{1}{n}\}^n$
\vspace{2pt}
\State $Z_{aug} \gets \{X_k^1\}_{k\in [1,n]}$
\vspace{2pt}
\For{$j \in [2,d]$}
\vspace{2pt}
    \State $Z_{aug}^{new} \gets \{\}$
    \vspace{3pt}
    \State $W_{aug}^{new} \gets \{\}$
    \vspace{3pt}
    \For{$Z_i, w_i \in Z_{aug}, W_{aug}$}
    \vspace{2pt}
        \For{$i_j \in [1,n]$}
            \State $w_i^{new} \gets w_i \cdot \frac{K^j(Z_i^{a(j)}-X_{i_j}^{a(j)})}{\sum_{k=1}^n K^j(Z_i^{a(j)}-X_k^{a(j)})}$
            \vspace{2pt}
            \State $Z_i^{new} \gets \{Z_i;X_{i_j}^{j}\}$
            \vspace{2pt}
            \If{$w_i^{new} > \theta$}
                \vspace{3pt}
                \State $Z_{aug}^{new} \gets Z_{aug}^{new} \cup Z_i^{new}$
                \vspace{3pt}
                \State $W_{aug}^{new} \gets W_{aug}^{new} \cup w_i^{new}$
            \EndIf
        \EndFor
    \EndFor
\EndFor
\State $Z_{aug} \gets Z_{aug}^{new}$
\vspace{3pt}
\State $W_{aug} \gets W_{aug}^{new}$
\end{algorithmic}
\textbf{Output:} $\hat{f} \in \text{arg}\,\min\limits_{f}\,\sum_{(w_i,Z_i)_{i \in (W_{aug},\,Z_{aug})}}w_iL(f,Z_i)$, $\;\; D_{aug}=(W_{aug}, Z_{aug})$
\end{algorithm}

\section{Experimental Design}\label{sec:setup}

\subsection{Dataset}

We relied on synthetic data to perform all the experiments. It enabled us to have full control of the problem represented by the data. Moreover, the simulated data were all sampled from~\glspl{scm}, since~\acrlong{ca} makes the assumption that the data are generated through a causal model. See~\citep{pearl:09} for a detailed definition of a~\gls{scm}\@.

We used the Causal Discovery Toolbox~\citep{kalainathan:20,kalainathan:22} to generate each~\gls{scm}. The~\gls{dag} of each~\gls{scm} was generated using the Erd\"{o}s-Rényi model~\citep{erdos:59} given a number of nodes and an expected degree. After each new edges' samples, we checked if it does not lead to cycle in the~\gls{dag}. The mechanism functions were generated from a set of parametric functions~(\eg{} linear or polynomial) whose parameters were randomly sampled from some given probability distributions, see~\cref{sec:cdt}. The source variables~(\ie{} vertices without parents in the~\acrlong{cg}) were generated using~\glspl{gmm} with four components and a spherical covariance. Finally, additive noise variables were introduced into the causal mechanisms. They were all i.i.d. and created according to a normal distribution\@.
Once a~\gls{scm} was built, the data were generated by sampling the realizations of the source and the noise variables. Finally, the mechanism functions computed the realizations of the variables following the topological order of the~\acrlong{cg}\@.

\subsection{Evaluation methodology}\label{subsec:methodology}

We considered different scenarios to assess the characteristics of ~\acrlong{ca} to provide some insights to practitioners about~\acrlong{ca}'s response to the various properties their problem might have. The scenarios, whose defaults parameters are detailed in~\cref{sec:default_param}, included:
\begin{itemize}
    \item \textbf{Non-linear data generation setting}: by varying the family functions of the mechanism included linear, polynomial, sigmoid, Gaussian process, and neural networks\@. 
    \item \textbf{Small-data regime}: by varying the number of observations from a few samples to a hundred samples~(\ie{} \([30, 40, 60, 80, 100, 300, 500, 700]\))
    \item \textbf{High-dimension scenario}: by varying the number of variables in a dataset from seven to twenty-five (\ie{} $[7, 8, 9, 10, 15, 20, 25]$)
    \item \textbf{Highly dependent input variables setting}: by varying the expected degree of the causal graph in $[0, 1, 2, 3, 4, 5, 6, 7]$
    \item \textbf{High aleatoric uncertainty setting}: by varying the additive noise amplitude in $[0.1, 0.2, 0.4, 0.6, 0.8, 1]$
    \item \textbf{Noisy acquisition procedure} (\ie{} outliers): by varying the fraction of outliers in $[0.01, 0.02, 0.03, 0.04, 0.05, 0.1, 0.15]$
    \item \textbf{Inadequate parametrization scenario}: by varying the probability threshold \(\theta\) defined in~\cref{sec:causalda}. \(\theta \in [10^{-1}, 10^{-2}, 10^{-3}, 10^{-4}, 10^{-5}]\)
\end{itemize}

For each of these scenarios, we compared the distributions of the original dataset and the augmented one by measuring the Kullback Leibler divergence, the Wasserstein distance, and the average relative difference in variance among the variables. %On the weighted augmented set generated by~\acrlong{ca}.

We also benchmarked~\acrlong{ca} against a baseline. In this case, we split the original dataset~(\(\mathcal{D}\)) into train and test sets following a \(70\%,30\%\) split strategy. Then, we trained two~\gls{xgb} models on the weighted augmented set~(\(D_{aug}:=({Z}_{i}, w_i), \; i\in[1,n^d]\)) and on the original training set to be our baseline. We measured their~\gls{mape} and R2 scores on the test set of the original dataset to predict each variable of the problem. 
Each~\gls{xgb} model was trained taking into account the data weights for the augmented set and uniform weights for the original training set) using a threefold cross-validation process to search from the best parameters set among the \(n\_\text{estimators}\in{}[10, 50, 200]\) and \(\text{reg}\_{\text{lambda}}\in{}[1, 10, 100]\)\@. 

For the outlier scenario, we additionally compare the distributions of the altered augmented data~(\ie{} with outliers) and the normally augmented data (\ie{} without outliers) by using the same metrics.

Finding the causal graph based on the observed variables is an NP-hard combinatorial optimization problem, which limits the scalability of existing approaches to a few dozen variables~\citep{chickering:96,chickering:04}. This is why we opted to start exploring this limitation in this work. Nevertheless, we will leave for future work the study in which the number of features is higher than the number of observations\@.

\section{Results}\label{sec:results}

This section describes the results of the scenarios described in~\cref{sec:setup}. \Cref{sec:mechanisms,sec:noise,sec:density} show complementary results, where one can see that~\acrlong{ca} and the baseline have similar performance when the input variables are highly correlated and that~\acrlong{ca} is independent of the aleatoric uncertainty of the data and the mechanisms of the underlying generation model.\@
\begin{figure}[!ht]
  \centering
  \footnotesize{%
  \begin{subfigure}[b]{0.49\linewidth}
     \includegraphics[width=\textwidth]{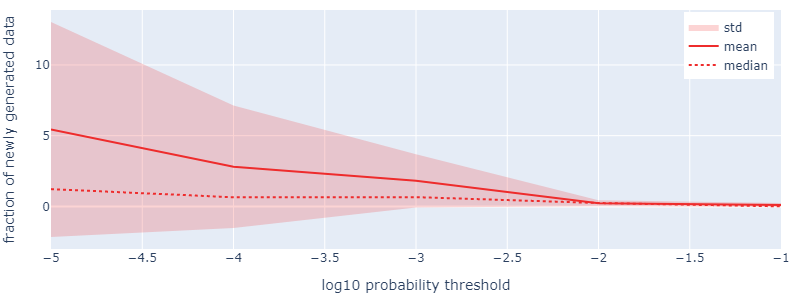}
     \caption{Fraction of new data}\label{fig:frac-new}
  \end{subfigure}
  \begin{subfigure}[b]{0.49\textwidth} 
     \includegraphics[width=\textwidth]{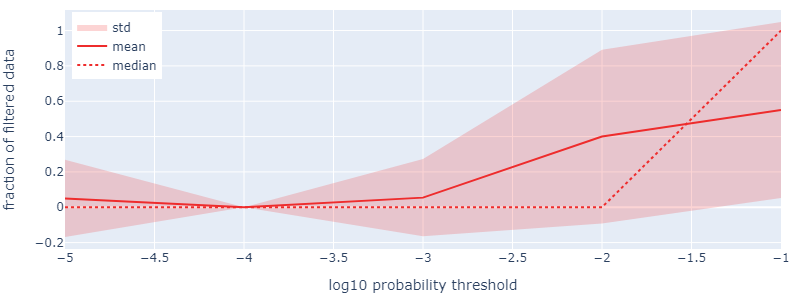}
     \caption{Fraction of filtered data}\label{fig:frac-filtered}
  \end{subfigure} \\
  \begin{subfigure}[b]{0.49\textwidth}
      \includegraphics[width=\textwidth]{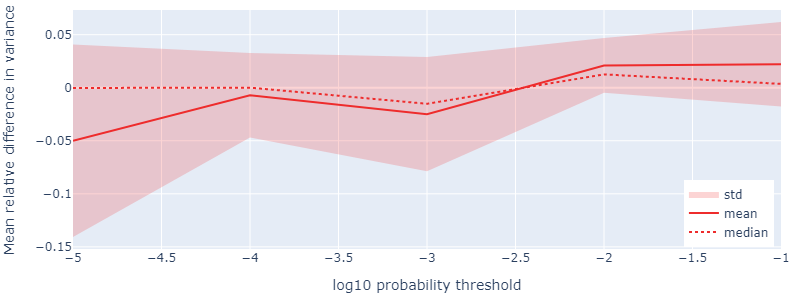}
      \caption{Relative difference in variance with augmentation}\label{fig:var-thr}
  \end{subfigure}
  \begin{subfigure}[b]{0.49\textwidth}
      \includegraphics[width=\textwidth]{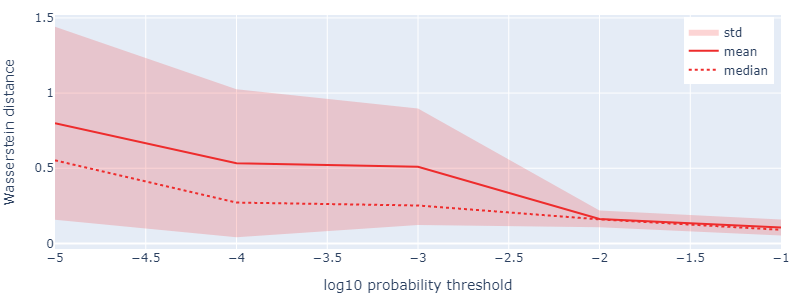}
      \caption{Wasserstein distance, original vs augmented sets}\label{fig:wss-orig-new}
  \end{subfigure}\\
  \begin{subfigure}[b]{\textwidth}
      \includegraphics[width=\textwidth]{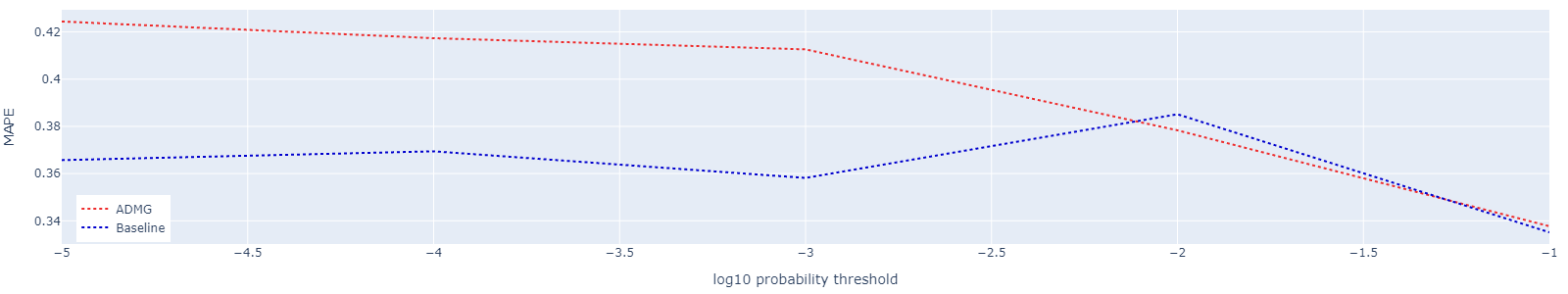}
      \caption{\gls{xgb} median MAPE score}\label{fig:threshold_res}
  \end{subfigure}
    \caption{~\acrlong{ca} output characteristics depending on the probability threshold}\label{fig:threshold_charac_results}
}
\end{figure}
\paragraph{Inadequate parametrization.} \Acrlong{ca} relies on its probability threshold parameter~\(\theta \in{}[0,1]\)~(\Cref{sec:causalda}) to prune the augmented data, which affects their distribution. While a probability threshold close to one accentuates the correlations of the observed data, a threshold close to zero relaxes them according to the~\acrlong{cg}, thus, generating more data points, as illustrated in~\cref{fig:frac-new,fig:frac-filtered,fig:wss-orig-new}. 
The fact that the variance decreases with the fraction of newly generated data,\cref{fig:var-thr} vs. \cref{fig:frac-new}, shows that~\acrlong{ca} does not tend to increase diversity in the dataset but changes its distribution in dense areas of the observed set. Hence, for an appropriate choice of probability threshold, we expect~\acrlong{ca} to improve an~\gls{ml} model predictions on the data support by providing a refined data distribution. \Cref{fig:threshold_res} illustrates this finding. 
The probability threshold parameter seems to have a very narrow value range to improve the performance of the~\gls{xgb} models. Thus, it must be carefully defined by the practitioners. 
% In addition to having a major effect on the~\gls{xbg}s' performance, the probability threshold improving predictions seems to have a very narrow range. 
%This highlights the fact that, in practice, practitioners' choice for this parameter is crucial. 
Indeed, this threshold can be interpreted as the minimal probability of accepting a new value for a variable given its parents. From~\cref{eq:weight}, one can easily see that \( {w}_{{i}} > \theta \implies w_i^j > \theta, \; \forall j \in [1,d]\). Hence, we encourage practitioners to analyze the distribution of each variable given its parents in order to make an informed choice on the probability threshold to use\@.

\begin{figure}%[!h]
  \centering
  \footnotesize{%
  \begin{subfigure}[b]{0.49\linewidth}
     \includegraphics[width=\textwidth]{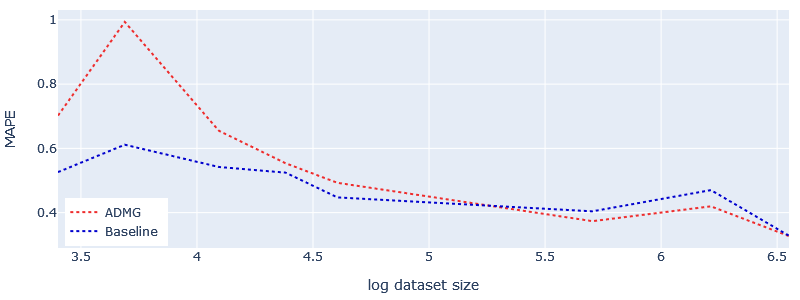}
     \caption{Median MAPE score}
  \end{subfigure}
  \begin{subfigure}[b]{0.49\textwidth} 
     \includegraphics[width=\textwidth]{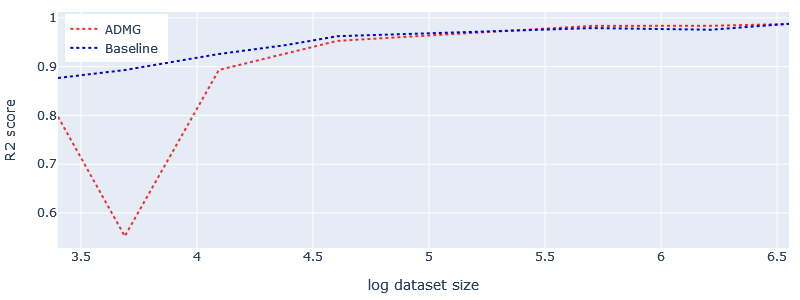}
     \caption{Median R2 score}
  \end{subfigure}
    \caption{\Gls{xgb} median performance depending on the size of the dataset}\label{fig:datasize_results}
}
\end{figure}

\begin{figure}%[!h]
  \centering
  \footnotesize{%
  \begin{subfigure}[b]{0.49\linewidth}
     \includegraphics[width=\textwidth]{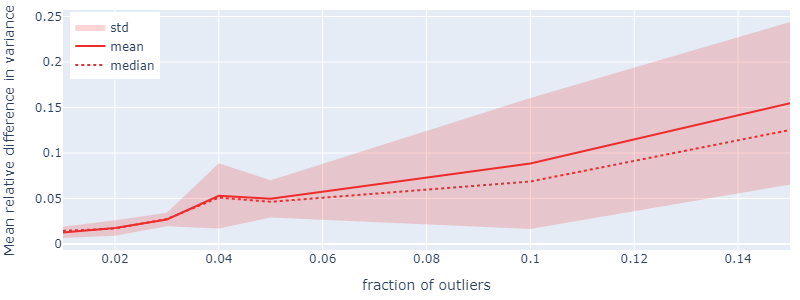}
     \caption{Relative difference in variance adding outliers}
  \end{subfigure}
  \begin{subfigure}[b]{0.49\textwidth} 
     \includegraphics[width=\textwidth]{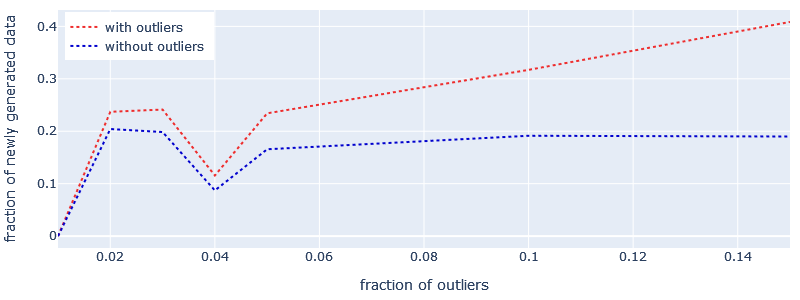}
     \caption{Median fraction of new data}
  \end{subfigure} \\
  \begin{subfigure}[b]{\textwidth}
      \includegraphics[width=\textwidth]{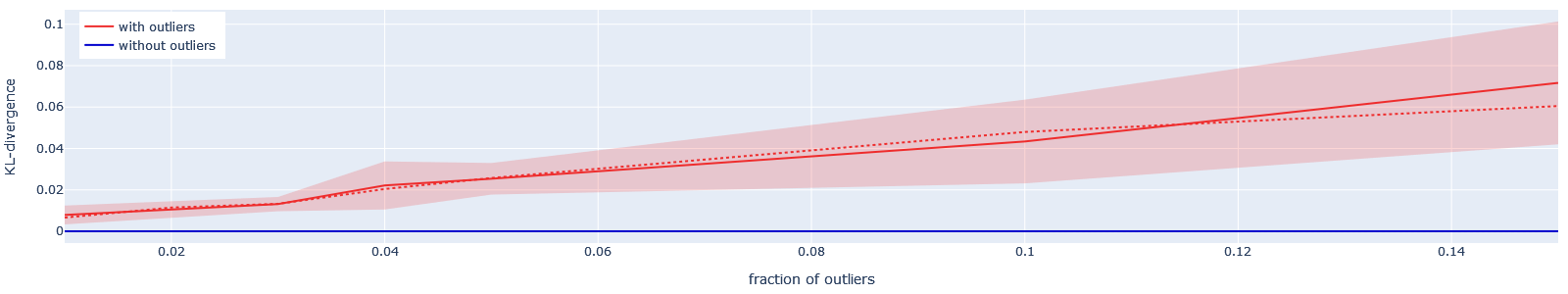}
      \caption{KL-divergence, corrupted vs normal sets, continuous line is mean, dashed line is median}
  \end{subfigure} \\
  \begin{subfigure}[b]{0.49\linewidth}
     \includegraphics[width=\textwidth]{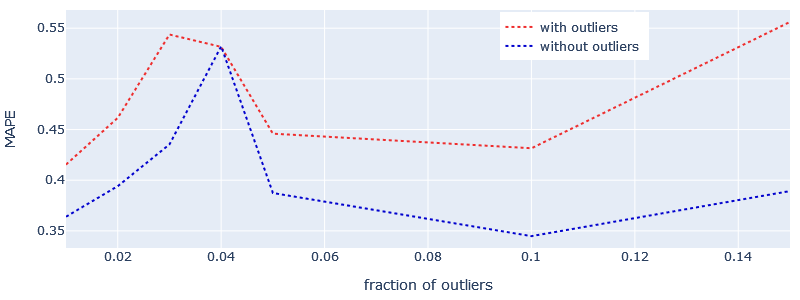}
     \caption{Madian~\gls{xgb}'s MAPE score}\label{fig:outliers_xgb_mape} 
  \end{subfigure}
  \begin{subfigure}[b]{0.49\textwidth} %fig:outliers_results
     \includegraphics[width=\textwidth]{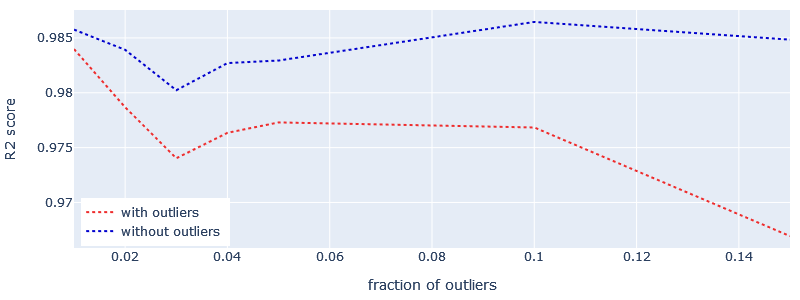}\label{fig:outliers_xgb_r2} 
     \caption{Median~\gls{xgb}'s R2 score}
  \end{subfigure}
    \caption{\acrlong{ca} performances under different level of outlier noise}\label{fig:outliers_data_results}
}
\end{figure}

% \begin{figure}[!h]
%   \centering
%   \footnotesize{%
%   \begin{subfigure}[c]{\textwidth} 
%      \includegraphics[width=\textwidth]{mape_threshold}
%   \end{subfigure}
%     \caption{XGB median MAPE score depending on the probability threshold}\label{fig:threshold_res}
% }
% \end{figure}
% 
\paragraph{Small-data regime.} \Cref{fig:datasize_results} shows that~\acrlong{ca} requires at least 300 observations to improve~\gls{xgb}'s performance. This quantity can be considered ``relatively high'' given the study scenario: \begin{inparaenum}[(a)] \item no outliers, \item use of the correct causal graph, \item in-distribution, \item data generated from~\gls{gmm}, and \item neural networks without high discontinuity or divergence\end{inparaenum}. 
%Hence, using~\acrlong{ca} in small-data regime may be challenging. 
Hence, improving some~\gls{ml} model predictions with~\acrlong{ca} in a small-data regime may be challenging. One can explain it by observing in~\cref{eq:weight} that the kernel density estimator overfits when there are only a few data points. Likewise, because each new data point is generated conditioned on the values taken by its parents, \acrlong{ca} needs several observed points with the same parents' realizations to generate new ones. %Second, and the most critical one, the kernel density estimates might have a high error due to a overfitted to the few available data points.
% 
% because we consider that the minimum amount of observations~(300) to improve XGB's results is high given that the current prediction problem is an easy one (no outliers, true causal graph, in-distribution, data generated from~\gls{gmm} and neural networks without high discontinuity or divergence). Hence, the method seems to not be suitable for small-data regime. 
% 
% This result is logical because, looking back at the equations, having a small amount of data creates two issues. First, because each new data point is generated conditioned on the values taken by its parents,~\acrlong{ca} needs several observed points with the same (or very close) parents' realization to generate new data. Second, and most critical one, kernel density estimates might have a high error because overfitted to the few available data points.
% 
Hence, we recommend considering~\acrlong{ca} not as a solution to compensate for the lack of data but rather as a method to refine the estimation of the data distribution via weighted data augmentation\@.
\begin{figure}[!ht]
  \centering
  \footnotesize{%
  \begin{subfigure}[b]{0.49\linewidth}
     \includegraphics[width=\textwidth]{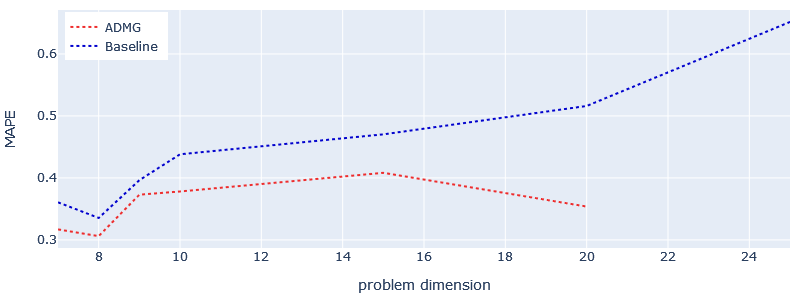}
     \caption{Median~\gls{xgb}'s MAPE score}
  \end{subfigure}
  \begin{subfigure}[b]{0.49\textwidth} 
     \includegraphics[width=\textwidth]{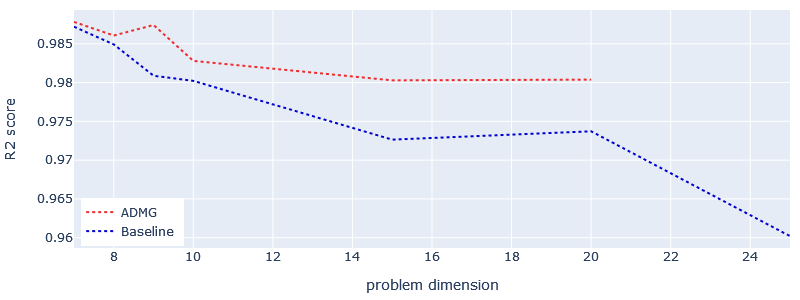}
     \caption{Median~\gls{xgb}'s R2 score}
  \end{subfigure} \\
  \begin{subfigure}[b]{\textwidth}
      \includegraphics[width=\textwidth]{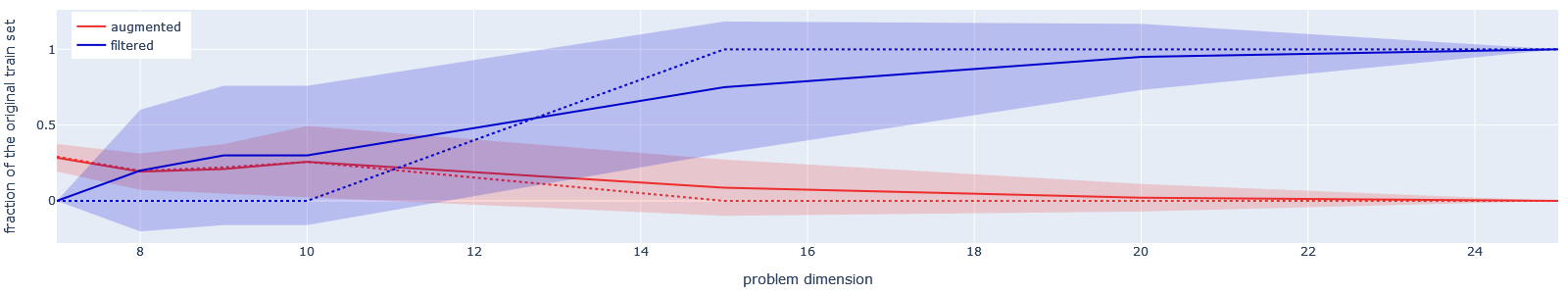}
      \caption{Fraction of new and filtered data points, continuous line is mean, dashed line is median}
  \end{subfigure}
    \caption{~\acrlong{ca} performances depending on problems dimension}\label{fig:dim_results}
}
\end{figure}
\paragraph{High-dimension scenario.} \Cref{fig:dim_results} shows that increasing dimension favors~\acrlong{ca}. Indeed,~\acrlong{ca} takes advantage of the prior knowledge about the conditional independence encoded in the~\acrlong{cg} to improve~\gls{ml} models' generalization. The literature has shown that such a problem is a challenge in low-dimensional data~(\cite{shah:20}) because high-dimensional settings with a known~\acrlong{cg} and expected degree lead to more conditional independence.
\Cref{fig:dim_results} also emphasizes that increasing the dimension increases the probability for the whole dataset to be filtered. Indeed, the probability threshold can be interpreted as follows: For a given \({Z}_{i}\), under the hypothesis that all the \(w_i^j\) are equal to \(c\), \(\sqrt[d]{\theta}\) is the minimum value of \(c\) for \({Z}_{i}\) not to be pruned. As \(\sqrt[d]{\theta}\) increases with \(d\), the higher the dimension, the higher the probability of the weights not to be pruned for a fixed probability threshold. As a result, practitioners also have to consider the dimension of the problem when choosing an appropriate probability threshold value\@.
\paragraph{Noisy acquisition procedure.} At first sight, we could expect~\acrlong{ca} to be robust to outliers thanks to its pruning strategy. Indeed, for a high enough probability threshold, the method can theoretically filter outliers with a low probability. However, even in the best-case scenario where all outliers are filtered, they are still taken into account by the Kernel density estimator, making them corrupted and spreading the effect of the outliers on the augmented set. Hence, the robustness of~\acrlong{ca} to outliers stays an open question\@.
From~\cref{fig:outliers_data_results}, it can be seen that~\acrlong{ca} propagates the outliers to the augmented set degrading~\gls{xgb}'s performance\@.
% 
% It uses the outliers to generate new data points thus propagating and increasing their influence resulting in a huge degradation on~\gls{xgb}' performance.

% \begin{figure}[!h]
%   \centering
%   \footnotesize{%
%   \begin{subfigure}[b]{0.49\linewidth}
%      \includegraphics[width=\textwidth]{mape_outliers}
%      \caption{Median MAPE score}
%   \end{subfigure}
%   \begin{subfigure}[b]{0.49\textwidth} 
%      \includegraphics[width=\textwidth]{r2_outliers}
%      \caption{Median R2 score}
%   \end{subfigure}
%     \caption{XGB performances under different level of outlier noise}\label{fig:outliers_results}
% }
% \end{figure}

% "due to its dependence to the aquision process" -> c'est pas plutot dans l'autre sens ?
\paragraph{Discussion.} The results presented in this section enabled us to understand under which conditions~\acrlong{ca} can help practitioners improve their~\gls{ml} models. First, we observed that \acrlong{ca} performance is independent of the underlying causal generation process. Nevertheless, it depends on the acquisition procedure because it is sensitive to outliers. Second, based on our experiments, the method requires at least 300 samples, making it unsuitable for small-data regimes. It can instead be used to improve the generalization of a~\gls{ml} model by providing a more refined data distribution, faithful to the observed one without increasing the diversity, using the prior knowledge encoded in the causal graph. Third,~\acrlong{ca} highly relies on the probability threshold parameter whose choice might be complex for practitioners. Indeed, up to now, no procedure has been developed to ensure a more guided choice for this parameter. We estimate that this last point is the most critical for practitioners to use this method in real-world use cases. That is why we would like to focus our future work on automating the choice of the probability threshold. Possible solutions include using the~\gls{ml} model to be trained to adjust this parameter automatically or employing~\gls{mcts}~\citep{kocsis:06} when computing the weights instead of the pruning procedure. 
%\Gls{mdl}~\citep{grunwald:07} can also be used to handle outliers by avoiding propagating them to the augmented dataset. In this case, one can compute a probabilistic score for each set of points previously identified\@.
Also, let us highlight that the experiments of this paper are not self-sufficient. Thus, we would like to deepen our evaluation by, first, using conditional independence tests to check if ~\acrlong{ca} indeed increases the conditional independences encoded in the causal graph, second, evaluating the method on real data, and third, analyzing the sensitivity to an erroneous causal graph. Indeed, building a causal graph might be challenging and, as far as we know, there is no general procedure to validate its truthful\@.

\section{Conclusion and Further Works}\label{sec:conclusion}

Data scarcity is a significant challenge when applying~\gls{ml} in high-stake domains such as healthcare and finance. Over the last few years, various approaches have been developed to enable researchers and practitioners to increase the size of their datasets artificially and, consequently, the robustness and generalization of their~\gls{ml} models. %Nevertheless, these methods mainly have targeted~\gls{cv} and~\gls{nlp} tasks. Furthermore, they rarely consider users' knowledge about the underlying data generation mechanism in a tabular data context, which often limits the fidelity and diversity of the generated data samples. 
Causal data augmentation strategies aim to handle these endeavors by relying on conditional independence encoded in a causal graph. 

This paper experimentally analyzed the~\acrlong{admg} data augmentation method~\citep{teshima:21} considering several scenarios. The goal was to help researchers and practitioners understand under which conditions their prior knowledge help in generating new data that enhance the performance of their models, as well as the influence of the parameters of the data augmentation strategy underneath the presence of outliers, error measures~(\ie{} aleatoric uncertainty), and the minimal number of samples of the observed data. Experimental results showed that the sample size is essential when employing the method. Likewise, it propagates the outliers when presented in the data. Furthermore, its hyperparameters must be carefully defined for each dataset. In future work, we plan first to carry out further experiments using, notably, conditional independence tests and real data and, secondly, to automatize the hyperparameters optimization process\@.

\section*{Acknowledgments}
\footnotesize{%
This research was partially funded by the European Commission within the HORIZON program~(TRUST-AI Project, Contract No. 952060)\@. 
}

%% Print the bibliography
\bibliographystyle{iclr2023}
\bibliography{references}

\begin{thebibliography}{20}
\providecommand{\natexlab}[1]{#1}
\providecommand{\url}[1]{\texttt{#1}}
\expandafter\ifx\csname urlstyle\endcsname\relax
  \providecommand{\doi}[1]{doi: #1}\else
  \providecommand{\doi}{doi: \begingroup \urlstyle{rm}\Url}\fi

\bibitem[Alaa et~al.(2022)Alaa, Van~Breugel, Saveliev, and van~der
  Schaar]{alaa:22}
Ahmed Alaa, Boris Van~Breugel, Evgeny~S Saveliev, and Mihaela van~der Schaar.
\newblock How faithful is your synthetic data? sample-level metrics for
  evaluating and auditing generative models.
\newblock In \emph{International Conference on Machine Learning}, pp.\
  290--306, 2022.

\bibitem[Chickering(1996)]{chickering:96}
David~Maxwell Chickering.
\newblock Learning bayesian networks is np-complete.
\newblock \emph{Learning from data: Artificial intelligence and statistics V},
  pp.\  121--130, 1996.

\bibitem[Chickering et~al.(2004)Chickering, Heckerman, and Meek]{chickering:04}
Max Chickering, David Heckerman, and Chris Meek.
\newblock Large-sample learning of bayesian networks is np-hard.
\newblock \emph{Journal of Machine Learning Research}, 5:\penalty0 1287--1330,
  2004.

\bibitem[Erdös \& Rényi(1959)Erdös and Rényi]{erdos:59}
P.~Erdös and A.~Rényi.
\newblock On random graphs i.
\newblock \emph{Publicationes Mathematicae Debrecen}, 6:\penalty0 290--297,
  1959.

\bibitem[Hao et~al.(2023)Hao, Zhu, Appalaraju, Zhang, Zhang, Li, and
  Li]{hao:23}
Xiaoshuai Hao, Yi~Zhu, Srikar Appalaraju, Aston Zhang, Wanqian Zhang, Bo~Li,
  and Mu~Li.
\newblock {MixGen}: A new multi-modal data augmentation.
\newblock In \emph{IEEE/CVF Winter Conference on Applications of Computer
  Vision}, pp.\  379--389, 2023.

\bibitem[Hendrycks et~al.(2021)Hendrycks, Basart, Mu, Kadavath, Wang, Dorundo,
  Desai, Zhu, Parajuli, Guo, et~al.]{hendrycks:21}
Dan Hendrycks, Steven Basart, Norman Mu, Saurav Kadavath, Frank Wang, Evan
  Dorundo, Rahul Desai, Tyler Zhu, Samyak Parajuli, Mike Guo, et~al.
\newblock The many faces of robustness: A critical analysis of
  out-of-distribution generalization.
\newblock In \emph{IEEE/CVF International Conference on Computer Vision}, pp.\
  8340--8349, 2021.

\bibitem[Ilse et~al.(2021)Ilse, Tomczak, and Forr{\'e}]{ilse:21}
Maximilian Ilse, Jakub~M Tomczak, and Patrick Forr{\'e}.
\newblock Selecting data augmentation for simulating interventions.
\newblock In \emph{International Conference on Machine Learning}, pp.\
  4555--4562, 2021.

\bibitem[Kalainathan et~al.(2020)Kalainathan, Goudet, and
  Dutta]{kalainathan:20}
Diviyan Kalainathan, Olivier Goudet, and Ritik Dutta.
\newblock {C}ausal {D}iscovery {T}oolbox: uncovering causal relationships in
  {P}ython.
\newblock \emph{The Journal of Machine Learning Research}, 21\penalty0
  (1):\penalty0 1406--1410, 2020.

\bibitem[Kalainathan et~al.(2022)Kalainathan, Goudet, Guyon, Lopez-Paz, and
  Sebag]{kalainathan:22}
Diviyan Kalainathan, Olivier Goudet, Isabelle Guyon, David Lopez-Paz, and
  Mich{\`e}le Sebag.
\newblock Structural agnostic modeling: Adversarial learning of causal graphs.
\newblock \emph{Journal of Machine Learning Research}, 23\penalty0
  (219):\penalty0 1--62, 2022.

\bibitem[Kocsis \& Szepesv{\'a}ri(2006)Kocsis and Szepesv{\'a}ri]{kocsis:06}
Levente Kocsis and Csaba Szepesv{\'a}ri.
\newblock Bandit based monte-carlo planning.
\newblock In \emph{17th European Conference on Machine Learning}, pp.\
  282--293, 2006.

\bibitem[Little \& Badawy(2019)Little and Badawy]{little:19}
Max~A Little and Reham Badawy.
\newblock Causal bootstrapping.
\newblock \emph{arXiv:1910.09648}, 2019.

\bibitem[Pearl(2009)]{pearl:09}
Judea Pearl.
\newblock Causal inference in statistics: An overview.
\newblock \emph{Statistics surveys}, 3:\penalty0 96--146, 2009.

\bibitem[Richardson(2003)]{richardson:03}
Thomas Richardson.
\newblock Markov properties for acyclic directed mixed graphs.
\newblock \emph{Scandinavian Journal of Statistics}, 30\penalty0 (1):\penalty0
  145--157, 2003.

\bibitem[Shah \& Peters(2020)Shah and Peters]{shah:20}
Rajen~D. Shah and Jonas Peters.
\newblock {The hardness of conditional independence testing and the generalised
  covariance measure}.
\newblock \emph{The Annals of Statistics}, 48\penalty0 (3):\penalty0 1514 --
  1538, 2020.

\bibitem[Talavera et~al.(2022)Talavera, Iglesias, González-Prieto, Mozo, and
  Gómez-Canaval]{talavera:22}
Edgar Talavera, Guillermo Iglesias, Ángel González-Prieto, Alberto Mozo, and
  Sandra Gómez-Canaval.
\newblock Data augmentation techniques in time series domain: a survey and
  taxonomy, 2022.

\bibitem[Teshima \& Sugiyama(2021)Teshima and Sugiyama]{teshima:21}
Takeshi Teshima and Masashi Sugiyama.
\newblock Incorporating causal graphical prior knowledge into predictive
  modeling via simple data augmentation.
\newblock In \emph{Uncertainty in Artificial Intelligence}, pp.\  86--96, 2021.

\bibitem[Van~Dyk \& Meng(2001)Van~Dyk and Meng]{van:01}
David~A Van~Dyk and Xiao-Li Meng.
\newblock The art of data augmentation.
\newblock \emph{Journal of Computational and Graphical Statistics}, 10\penalty0
  (1):\penalty0 1--50, 2001.

\bibitem[Wen et~al.(2021)Wen, Sun, Yang, Song, Gao, Wang, and Xu]{qingsong:21}
Qingsong Wen, Liang Sun, Fan Yang, Xiaomin Song, Jingkun Gao, Xue Wang, and
  Huan Xu.
\newblock Time series data augmentation for deep learning: A survey.
\newblock In \emph{Thirtieth International Joint Conference on Artificial
  Intelligence}, 2021.

\bibitem[Xie et~al.(2020)Xie, Dai, Hovy, Luong, and Le]{xie:20}
Qizhe Xie, Zihang Dai, Eduard Hovy, Thang Luong, and Quoc Le.
\newblock Unsupervised data augmentation for consistency training.
\newblock \emph{Advances in neural information processing systems},
  33:\penalty0 6256--6268, 2020.

\bibitem[Zhong et~al.(2020)Zhong, Zheng, Kang, Li, and Yang]{zhong:20}
Zhun Zhong, Liang Zheng, Guoliang Kang, Shaozi Li, and Yi~Yang.
\newblock Random erasing data augmentation.
\newblock \emph{AAAI Conference on Artificial Intelligence}, 34\penalty0
  (07):\penalty0 13001--13008, 2020.

\end{thebibliography}

\appendix
\section{Appendix}\label{sec:appendix}

\subsection{Non-linear data generation scenario -- results}\label{sec:mechanisms}

\begin{figure}[!ht]
  \centering
  \footnotesize{%
  \begin{subfigure}[b]{0.49\linewidth}
     \includegraphics[width=\textwidth]{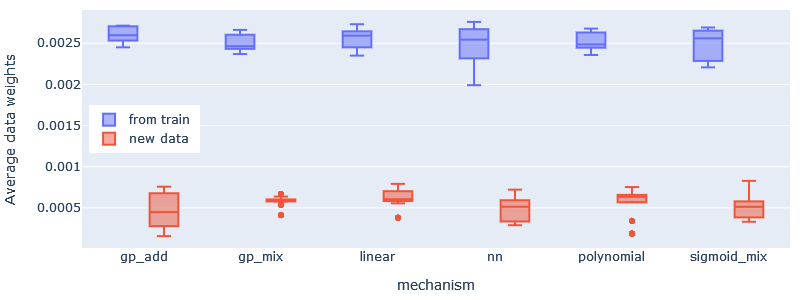}
     \caption{Average weights of the augmented dataset}
  \end{subfigure}
  \begin{subfigure}[b]{0.49\textwidth} 
     \includegraphics[width=\textwidth]{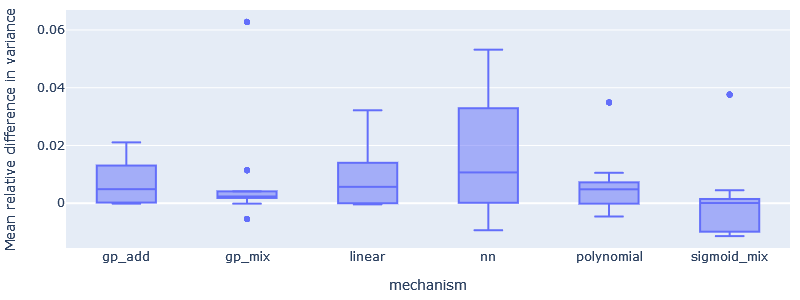}
     \caption{Relative difference in variance with augmentation}
  \end{subfigure} \\
  \begin{subfigure}[b]{0.49\textwidth}
      \includegraphics[width=\textwidth]{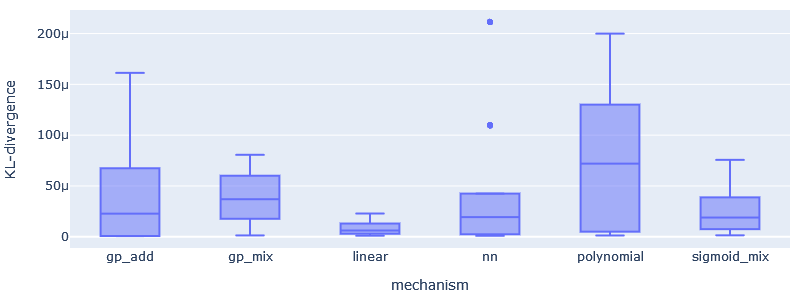}
      \caption{KL-divergence, augmented vs original sets}
  \end{subfigure}
  \begin{subfigure}[b]{0.49\textwidth}
      \includegraphics[width=\textwidth]{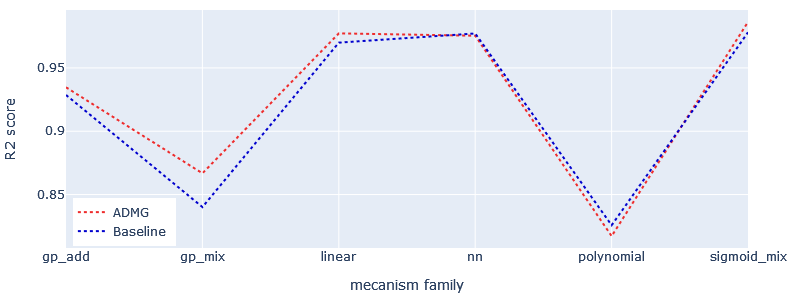}
      \caption{Median~\gls{xgb}'s R2 score}
  \end{subfigure}
    \caption{~\acrlong{ca} performance under different underlying mechanisms}\label{fig:mechanisms_results}
}
\end{figure}

As~\acrlong{ca} only computes the densities, see~\cref{eq:weight}, the type of mechanisms linking the variables should not affect the performances of the method, which is illustrated by~\cref{fig:mechanisms_results}. Indeed, what matters the most is whether a variable is continuous or discrete because it will affect the choice of the Kernel to use. Moreover, each practitioner can decide to choose a different Kernel based on the distribution of the variables given their parents. A common choice is to use a Gaussian Kernel with a Silverman bandwidth for continuous variables and the identity Kernel for the discrete ones.

Hence, \acrlong{ca} can be used in non-linear settings without special care.

\subsection{Highly dependent input variables scenario -- results}\label{sec:density}

\begin{figure}[!ht]
  \centering
  \footnotesize{%
  \begin{subfigure}[b]{0.49\linewidth}
     \includegraphics[width=\textwidth]{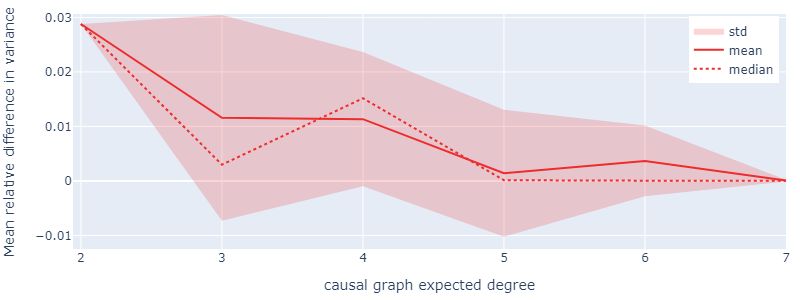}
     \caption{Relative difference in variance with augmentation}
  \end{subfigure}
  \begin{subfigure}[b]{0.49\textwidth} 
     \includegraphics[width=\textwidth]{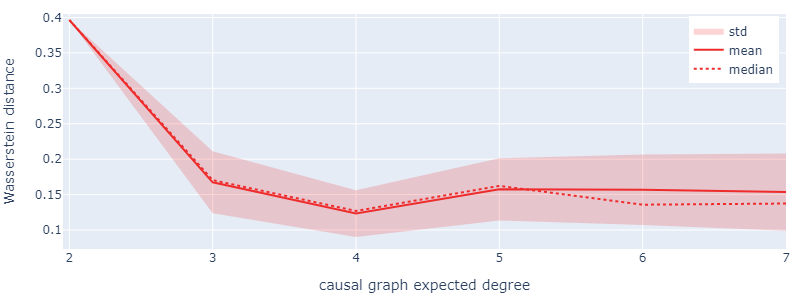}
     \caption{Wasserstein distance of the augmented vs. original sets}
  \end{subfigure} \\
  \begin{subfigure}[b]{\textwidth}
      \includegraphics[width=\textwidth]{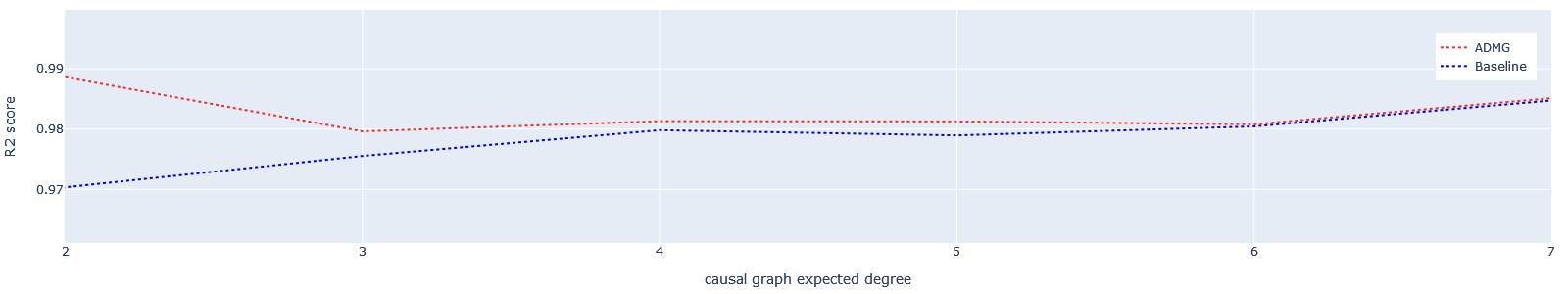}
      \caption{Median~\gls{xgb}'s R2 score}
  \end{subfigure}
    \caption{~\acrlong{ca} performance under different levels of connectivity in the~\gls{cg}}\label{fig:density_results}
}
\end{figure}

From~\cref{fig:density_results}, it can be observed that the more dependent the input variables, the closer to the baseline~\acrlong{ca}. Looking at the equations from~\cref{sec:causalda}, this is logical. Indeed, increasing the~\acrlong{cg} density implies that, on average, the number of parents also increases. As a result, the densities are computed on higher dimension supports, making them much smaller and more diluted, which does not encourage the generation of new samples. 

In other words, as~\acrlong{ca} aims to use the conditional independencie induced by the causal graph; if the number of edges increases, the number of conditional independence decreases, making thus the method less useful because it has less prior knowledge to use.

\subsection{Aleatoric uncertainty -- results}\label{sec:noise}

\begin{figure}[!ht]
  \centering
  \footnotesize{%
  \begin{subfigure}[b]{0.49\linewidth}
     \includegraphics[width=\textwidth]{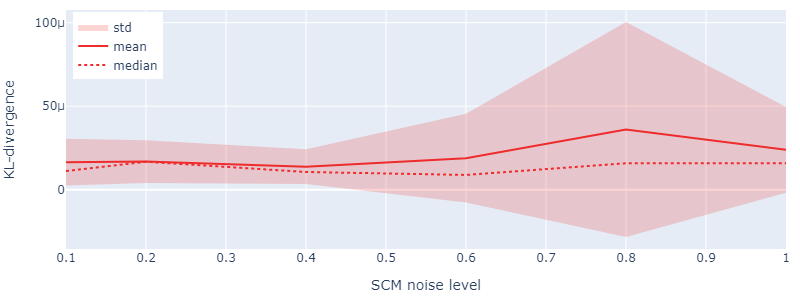}
     \caption{KL-divergence, augmented vs original sets}
  \end{subfigure}
  \begin{subfigure}[b]{0.49\textwidth} 
     \includegraphics[width=\textwidth]{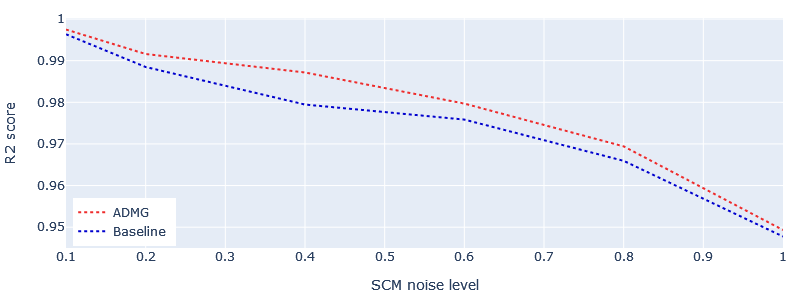}
     \caption{Median~\gls{xgb}'s R2 score}
  \end{subfigure} \\
  \begin{subfigure}[b]{\textwidth}
      \includegraphics[width=\textwidth]{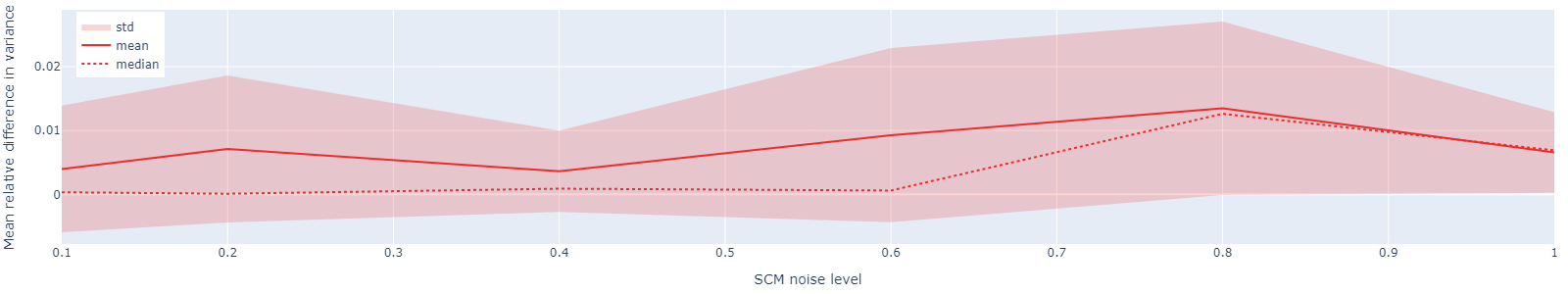}
      \caption{Mean relative difference in variance with augmentation}
  \end{subfigure}
    \caption{~\acrlong{ca} performance under different level of aleatoric uncertainty}\label{fig:noise_results}
}
\end{figure}

Based on the results from~\cref{fig:noise_results}, it can be asserted that the amplitude of the noise introduced in the~\gls{scm} generating the data does not have a significant influence on the results of~\acrlong{ca}. This makes sense. Indeed, varying the amplitude of the~\gls{scm}s' noise will only have an effect of scale on the density distributions which could easily be compensated by the bandwidth of the Gaussian kernels automatically optimized with the Silverman formula.

\subsection{Causal Discovery Toolbox -- \textit{AcyclicGraphGenerator} module description}\label{sec:cdt}

Causal Discovery Toolbox~\citep{kalainathan:20,kalainathan:22} is a Python package for causal inference. It mainly focuses on causal discovery in the observational setting. However, this section is dedicated to the description of the \textit{AcyclicGraphGenerator} module we used for our experiments.

The \textit{AcyclicGraphGenerator} module is able to randomly generate a~\gls{scm} associated with a dataset. The generator provides the user the ability to choose which mechanism to be used in the data generation process as well as the type of noise contribution (additive and/or multiplicative). 

Currently, the implemented mechanisms are:
\begin{itemize}
    \item \textbf{Linear}: \(y = \mathbf{X}W +\times E\)
    \begin{itemize}
        \item \(W \sim {\mathcal{U}[0,1]}^{D}\)
    \end{itemize}
    \item \textbf{Polynomial}: \(y = \sum_{i=0}^{d}\mathbf{X}^{i}W_i +\times E\)
    \begin{itemize}
        \item \(d\) the degree
        \item \(W_i \sim {\mathcal{U}[0,1]}^{D} \;\; \forall i\)
    \end{itemize}
    \item \textbf{Gaussian Process}: \(y = \sum_{i=0}^{D}s_i +\times E\) with  and 
    \begin{itemize}
        \item \(D\) the number of causes
        \item \(s_i \sim \mathcal{N}(0,cov(\mathbf{X}_i)) \;\; \forall i\)
    \end{itemize}
    \item \textbf{Sigmoid}: \(y = \sum_{i=1}^{D} (1+a) \cdot \frac{b \cdot (\mathbf{X}_{i}+c)}{1+|b \cdot (\mathbf{X}_{i}+c)|} +\times E\)
    \begin{itemize}
        \item \(D\) the number of causes
        \item \(a \sim Exp(4)\)
        \item \(b \sim \tilde{b} \cdot \mathcal{U}[-2,-0.5] + (1-\tilde{b}) \cdot \mathcal{U}[0.5,2], \; \tilde{b} \sim Ber(0.5)\)
        \item \(c \sim \mathcal{U}[-2,2]\)
    \end{itemize}
    \item \textbf{Randomly initialized Neural network}: \(y = \sigma((\mathbf{X}, E)W_{in})W_{out} \)
    \begin{itemize}
        \item \(\sigma\) the hyperbolic-tangent activation function
        \item \(W_{in}\) and \(W_{out}\) randomly initialized with the Glorot uniform
    \end{itemize}
\end{itemize}
with \(+\times\) denoting either addition or multiplication, \(\mathbf{X}\) the vector of causes of dimension \(D\), and \(E\) the noise variable accounting for all unobserved variables. As mentioned in~\cref{sec:setup}, \(E \sim \mathcal{N}(0,1)\) in our experiments.

To generate a random~\gls{scm} associated with a dataset, one needs to specify:
\begin{itemize}
    \item the functions family of the mechanisms
    \item the type of noise to use in the generative process (either ``uniform'' for a \(\mathcal{U}[-2,2]\) or ``gaussian'' for a \(\mathcal{N}(0,1)\) )
    \item the proportion of noise in the mechanism
    \item the number of observations to generate
    \item the number of nodes/variables in the~\acrlong{scm}
    \item the type of~\gls{dag} to generate (either 'default' to be sampled from the default procedure or ``erdos'' to be sampled from the Erd\"{o}s-Rényi model~\citep{erdos:59} augmented with a conditioned on the new sampled edges to check if it does not lead to a cycle)
    \begin{itemize}
        \item if ``default``: a maximum number of parents per node has to be specified
        \item if ``erdos'': an expected degree for the~\gls{dag} has to be specified
    \end{itemize}
\end{itemize}
Then, each~\gls{scm} is generated according to the following procedures:
\begin{enumerate}
    \item The~\gls{dag} is generated 
    \item The mechanism functions are generated 
    \item The source variables~(\ie{} vertices without a parent in the causal graph) are generated using~\glspl{gmm} with four components and with a spherical covariance\@.
    \item Noise variables are introduced into the causal mechanisms. They are all i.i.d.
\end{enumerate}

Once a~\gls{scm} is built, the data are generated by sampling the realizations of the source and the noise variables. Next, the mechanism functions compute the realizations of the variables following the topological order of the~\gls{dag}.

\subsection{Experiments' default parameters}\label{sec:default_param}

\begin{table}[!ht]
    \centering\footnotesize{%
    \caption{Parameters' values when not under study}\label{tab:default-parameters}%
    \begin{tabularx}{\linewidth}{rX}        
        \toprule
        \multicolumn{1}{c}{\textbf{Parameter}} & \multicolumn{1}{c}{\textbf{Value}}\\
        \midrule
        \multirow{2}{*}{\textbf{Network architecture}} & 2-layers fully-connected neural network with hyperbolic tangent activation function and 20 neurons initialized through the Glorot uniform \\
        \rowcolor{gray!25}
        \textbf{Number of variables} & 10 \\ %
         \textbf{Causal graph expected degree} & 3\\
         \rowcolor{gray!25}
         \textbf{Additive noise amplitude} & 0.4 \\
         \textbf{Probability threshold} & $10^{-2}$ \\
         \rowcolor{gray!25}
         \textbf{Fraction of outliers} & 0 \\
         \textbf{Number of repetitions} & 20\\
         \rowcolor{gray!25}
         \textbf{Kernels function} & Gaussian Kernels with Silverman bandwidth \\
        \bottomrule
    \end{tabularx}%
}
\end{table}

\end{document}